\documentclass[a4]{article}

\usepackage{graphicx}
\usepackage{caption}

\usepackage{longtable}
\usepackage{multirow}

\usepackage{booktabs}

\title{The Influence of Multiple Classes on Learning Online Classifiers from Imbalanced and Concept Drifting Data Streams}

\author{Agnieszka Lipska \\
Poznan University of Technology \\
{\tt agnieszka.l1995@gmail.com}
 \and Jerzy Stefanowski \\
 Institute of Computing Science, Poznan University of Technology,  \\ 60-965 Pozna\'n, Poland \\
 {\tt jerzy.stefanowski@cs.put.poznan.pl}
 } 
 
 \date{}

\begin{document}

\maketitle

\begin{abstract}
This work is aimed at the experimental studying the influence of  local data characteristics and drifts on the difficulties of learning various online classifiers from multi-class imbalanced data streams. Firstly we present a categorization of these data factors and drifts in the context of imbalanced streams, then we introduce the generators of synthetic streams that model these factors and drifts.  The results of many experiments with synthetically generated data streams have shown a much greater role of the overlapping between many minority classes (the type of borderline examples)  than for streams with one minority class. The presence of rare examples in the stream is the most difficult single factor. The local drift of splitting minority classes is the third influential factor. Unlike binary streams, the specialized UOB and OOB classifiers perform well enough for even high imbalance ratios. The most challenging for all classifiers are complex scenarios integrating the drifts of the identified factors  simultaneously, which worsen the evaluation measures in the case of a several minority classes stronger than for binary ones.
\noindent This is an extended version of the short paper presented at LIDTA'2022 workshop at ECMLPKDD2022 \cite{Lipska2022}.
\end{abstract}

\section{Introduction}
\label{sec:intro}

Many methods for improving learning classifiers from the imbalanced data have been developed in the last twenty  years, for surveys see, e.g., \cite{Branco,imb-chapter-book}. Nevertheless most of these researches concern static data that is fully accessible and  can be processed multiple times. In many modern data-intensive applications massive volumes of data are continuously generated in the form of \textit{data streams}.  Besides new computational requirements, another challenge for mining streams is their non-stationary characteristics, where the data and target concepts change over time in a phenomenon of \textit{concept drift}~\cite{ConceptAdaptationSurvey}. It poses needs for developing new, specialized  algorithms~\cite{DataStreamsOpenChallenges,krawczyk2017}. Learning classifiers in the context of non-stationary data streams has been intensively studied in the last twenty years ~\cite{PolikarSurvey,gamaBook2010,krawczyk2017}. It results in drift taxonomies~\cite{gamaBook2010}, special  detectors~\cite{ConceptAdaptationSurvey}, and many proposals of adaptive streaming classifiers~\cite{BrzezStefIEEE2013,PolikarSurvey,krawczyk2017}.


The task of  learning classifiers from streams becomes even more challenging in the presence of additional data complexities, in particular \textit{imbalances between cardinalities of target classes}. However, the current research on imbalanced and concept drifting streams are not as developed as in the case of separately considered static data or streams. The number of new algorithms for dealing with imbalanced stream classification is still much smaller \cite{Korycki2021,krawczyk2017}. Moreover, existing works mostly focus on re-balancing classes and reacting to changes affecting the global imbalance ratio.  These works do not sufficiently consider more  complex imbalanced streams scenarios, where these changes are accompanied by  the \textit{local difficulty factors} already considered for the static imbalanced data such as:  the fragmentation of the minority class  into sub-concepts, class overlapping or occurrence of different types of unsafe minority examples (borderline, rare or outliers \cite{Krysia_2016}) inside the distribution of the examples from particular classes. 

These data difficulty factors have already been investigated for static data and proved to be useful for understanding sources of difficulties in imbalanced classification problems or developing new algorithms, see e.g. \cite{krawczyk2016learning,Lopez2013,Stefanowski_2015}. However in evolving data streams they could also influence the changes in \textit{local class distributions} and other \textit{local drifts}.  Nevertheless,  the conjunctions of these data factors and drifts have not been sufficiently studied yet, see discussions in \cite{aguiar2022survey,BrzezStef2019,brzezinski2021impact}. 



A new categorization of concept drifts with data difficulty factors for class imbalanced streams is introduced in \cite{brzezinski2021impact}. These  authors also carried out comprehensive experiments with synthetic and real data streams, where they deeply studied the influence of concept drifts, global class imbalance, local data difficulty factors, and their combinations, on predictions of representative  online classifiers, such as online bagging  ensemble \cite{Oza}, VFDT classification tree \cite{vfdt},  two extensions of  online bagging ensembles designed to deal with imbalanced streams: OOB and UOB,  \cite{WangMY15,MUOB},  and ESOS-ELM neural network ensemble \cite{ESOS}. Experimental results showed differences in existing classifiers’ reactions to such factors.   Combinations of multiple factors demonstrated to be the most challenging for classifiers, which were not able to recover from these drifts.  For instance, the specialized imbalanced stream classifiers, such as UOB and OOB, were able to cope well with global class imbalance (except the highest ones), whereas non-specialized classifiers (e.g. VFDT) performed worse.

Nevertheless  \cite{brzezinski2021impact} is limited to considering binary imbalanced classes only. In the current paper we carry out  new experiments with studying the aforementioned data factors and drifts for \textit{multiple classes streams}, which has not been investigated yet in the current literature on imbalanced data streams.

Recall that multiple imbalanced classes are often identified as more difficult than binary data in case of static data \cite{MUOB}. We want also verify observations from the study \cite{brzezinski2021impact}  will be valid also for the multiple classes. Furthermore following other experiments with static multiple data, such as the recent study \cite{eswa2022}, we are interested to check whether the role of borderline examples will be more important in case of many overlapping classes. Therefore,  we decided to create an additional generator of synthetic streams. Unlike experimental studies aimed at comparisons of several classifiers, we claim that this kind of study could support  better understanding the difficulties of this type of drifting data streams and at the same time could indicate the directions of development of new specialized algorithms.

The paper is organized as follows. In section \ref{sec:categorization} we present our categorization of data difficulty factors and local concept drifts. Then,  in the next section we introduce our experimental setup and characterize scenarios of generating synthetic data sets. It is followed by the presentation of main results from our experimental study. The final section groups conclusions and discusses future research.

\section{Categorization of data difficulty factors and local concept drifts}
\label{sec:categorization}

A data stream is a sequence of labeled examples $\{\mathbf{x}^t, y^t\}$ for $t = 1, 2, \ldots, T$, where $\mathbf{x}$ is a vector of attribute values and $y$ is a class label ($y \in \{K_1,\ldots, K_l\ $). It could be processed either \textit{online} or in \textit{blocks} (\textit{data chunks}). Alongside restrictions concerning processing time and memory usage data streams are characterized by their potential changes over time, i.e. the  \textit{concept drift}~\cite{ConceptAdaptationSurvey}. More formally, let for each point in time $t$, every example is generated by a source with a joint distribution $P^t(\mathbf{x},y)$.  The concept drift occurs if  $\mathbf{x}$  $P^t(\mathbf{x},y) \neq P^{t+\Delta}(\mathbf{x},y)$ holds for for two distinct points in time $t$ and $t+\Delta$ \cite{gamaBook2010}.  Depending on the rate and severity concept drifts are categorized into sudden, gradual, incremental and recurrent ones \cite{ConceptAdaptationSurvey}.  

Up to know several classifiers specialized for concept drifting streams have been proposed \cite{PolikarSurvey,gamaBook2010,krawczyk2017}. 
They can be divided into \textit{active} (trigger-based) approaches, which introduce changes in classifiers when drifts are detected, and \textit{passive} (adaptive) approaches, which continuously update the classifier, in particular streaming ensembles, see e.g. \cite{krawczyk2017}.

The imbalance between classes is characterized by the \textit{global imbalance ratio}. Here for multiple classes, these ratios are defined as the percentage of examples in the data that belong to the minority class \cite{eswa2022}. In case of static imbalanced data it has been shown that beside the global imbalance ratio that other sources  of classifiers deterioration include: (1) the decomposition of the minority class into several sub-concepts, (2) the presence of small, isolated groups of minority examples located deeply inside the majority class region (it corresponds to \textit{rare cases}), (3) the effect of strong overlapping between the classes, see for more comprehensive presentations e.g. \cite{Sanchez,JoJ04,krawczyk2016learning,Lopez2013,Prati,Stefanowski_2015,santos2022joint}. 

The last two factors can be identified through the so-called \textit{types of examples} \cite{NapieralaS12,Krysia_2016}, which distinguished between safe and unsafe examples. \textit{Safe examples} are the ones located in homogeneous regions populated by examples from one class only. The \textit{unsafe}  examples are categorized into \textit{borderline}~(placed close to the decision boundary between classes), \textit{rare cases}~(isolated groups of few examples located deeper inside the opposite class), and \textit{outliers}.  Following the method from \cite{Krysia_2016} these types of examples can be identified based on the analysis of class labels of other examples in the local neighborhood. Although they were initially defined for binary imbalanced data, they were furthered generalized for multiple imbalanced classes \cite{LNS2017} and recently exploited in a comprehensive experimental study of difficulties of static multi-class imbalanced data sets \cite{eswa2022}. 

Most of the aforementioned studies on imbalanced data streams focus on handling either static or drifting imbalance ratios only and they do not consider  the aforementioned data factors.

The recent work \cite{brzezinski2021impact} introduced an \textit{extended concept drift categorization from imbalanced streams}  that takes into account these local data factors into account -- see Table \ref{drifty}, which covers  four main criteria and specific types of drifts inside them.  Its important characteristic is distinguishing \textit{locality in the data streams}, i.e. the interest  is focused on the local bounded sub-regions in the attribute space and changes occuring in these local regions as opposed to global drifts.

\begin{table}
\begin{center}
\caption{The imbalanced stream drift categorization, following \cite{brzezinski2021impact} }
\label{drifty}
\begin{tabular}{|c|c|}
 \toprule
Criteria & Drifts \\
\midrule
Drift region & local region drift \\ 
 global vs local & global drift \\
  & global-local drift \\ \hline
Class composition & class split drift \\
 homogeneous vs. heterogeneous & class merge drift \\
 & split and merge drifts \\  \hline
 Imbalanced ratio & dynamic imbalanced ratio \\
 Balanced vs. imbalanced & class swap \\ \hline
 Distribution of minority class & minority example type \\
 Safe, Borderline, Rare, Outlier & distribution change \\
\bottomrule
\end{tabular}
\end{center}
\end{table}

The first criterion corresponds to distinguishing between types of drifts, where the probability distribution associated with the whole attribute space or its local parts.  A \textit{homogeneous} class composition means that examples of the minority class are concentrated in a single local region  A \textit{heterogeneous} class composition means that examples of the class are spread over multiple local regions (which are larger than groups of rare minority examples). It corresponds to mainly drifts being class region split or merging them.  Then, two situations are considered: \textit{static imbalance ratio} if the class proportions do not change over time, and \textit{dynamic imbalance ratio} if the imbalance ratio changes over time. The last criterion is referred to monitoring proportion of the types of the minority class examples in the stream, which could be changing or static over streams.  For more details see \cite{brzezinski2021impact}.

\section{An experimental setup}

The aim of experiments is to investigate the influence of the data difficulty factors and drifts on predictive performance of selected online classifiers. As these experiments refer to the earlier study with binary imbalanced streams, we follow and extend its experimental setup \cite{brzezinski2021impact}. In general we want to answer two research questions: 
\begin{itemize}
\item[\textbf{RQ1}] What is the impact of different types of single data difficulty factors and isolated, single drifts on the classification performance? 
\item[\textbf{RQ2}] Which complex scenarios integrating several data factors and drifts are the most harmful for classification performance?  In this context we want also compare new results for multiple imbalanced classes with earlier ones obtained by binary (minority vs. majority) classes \cite{brzezinski2021impact}.
\end{itemize}

\noindent Our experiments cover the following difficulty factors and drifts:
\begin{itemize}
 \item \textit{Imbalanced ratios} -- into two scenarios of either  static data or with a single drift; We consider the following configurations: one majority class and two minority classes with ratios (sizes referred to all examples) (1\% 1\%); (3\% 3\%); (5\% 5\%); (10\% 10\%) and the balanced scenario (30\% 30\%); and with three minority classes (1\% 1\% 1\%); (3\% 3\% 3\%); (5\% 5\% 5\%); (10\% 10\% 10\%) and the balanced version with 25\%. 
 \item \textit{Types of minority examples} being either \textit{borderline} (class overlap) or \textit{rare} ones (other examples in the minority  classes are generated to be safe ones), i.e. with the given percentages of their occurrence in each class: 20\%, 40\%, 60\%, 80\% and 100\%.
 \item \textit{Changes in class composition}, i.e. local drifts of the split of the each minority class into 3, 5 and 7 sub-clusters; and moving these numbers of sub-clusters. We omitted scenarios with merging sub-clusters into the single concepts as they were not influential in experiments of \cite{brzezinski2021impact}.
 \end{itemize} 

We carried out the experiments in a controlled framework based on synthetic generated data, where each data factor can be modeled and parametrized according to different planned scenarios.  We prepared two generators: 
\begin{enumerate}
\item The \textit{old generator} - the same as used in the previous experiments \cite{brzezinski2021impact} where minority classes are generated in elliptical spheres and the majority class instances uniformly surround them. Its first version coded by \cite{pewinski} has been then adapted by \cite{lipska} to produce streams for multiple classes.  Its description is available as an appendix to \cite{brzezinski2021impact}. The source code of this implementation  for binary classes is available at:\\  https://github.com/dabrze/imbalanced-stream-generator.
\item The \textit{new generator} -- the majority class has an elliptical shape and its overlaps with the remaining class of also the similar shape. It is also possible to model Gaussian distribution in each class region. Its detailed description is given in \cite{lipska}.
\end{enumerate}

A more detailed description of both generators with some graphical illustration is provided in an additional appendix to the main paper, see \cite{appendixLipska}.

\begin{figure}[h]
 \centering
    \begin{minipage}{0.45\textwidth}
        \centering
        \includegraphics[width=0.9\textwidth]{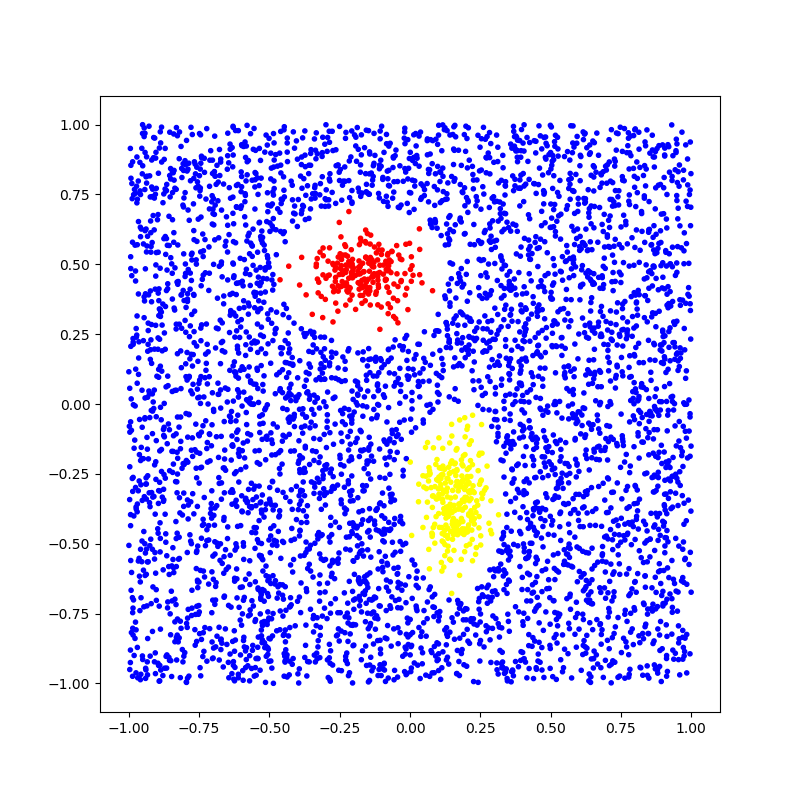} 
    \end{minipage}
    \caption{Two dimension visualization of examples from three classes generated by the old generator. Two minority classes are located in the center of the attribute space. Red and yellow color points represent safe examples in each class, surrounded by white border examples. Blue color points show examples from the majority class.}
\label{oldgen}
\end{figure}

\begin{figure}[h]
 \centering
    \begin{minipage}{0.45\textwidth}
        \centering
        \includegraphics[width=0.9\textwidth]{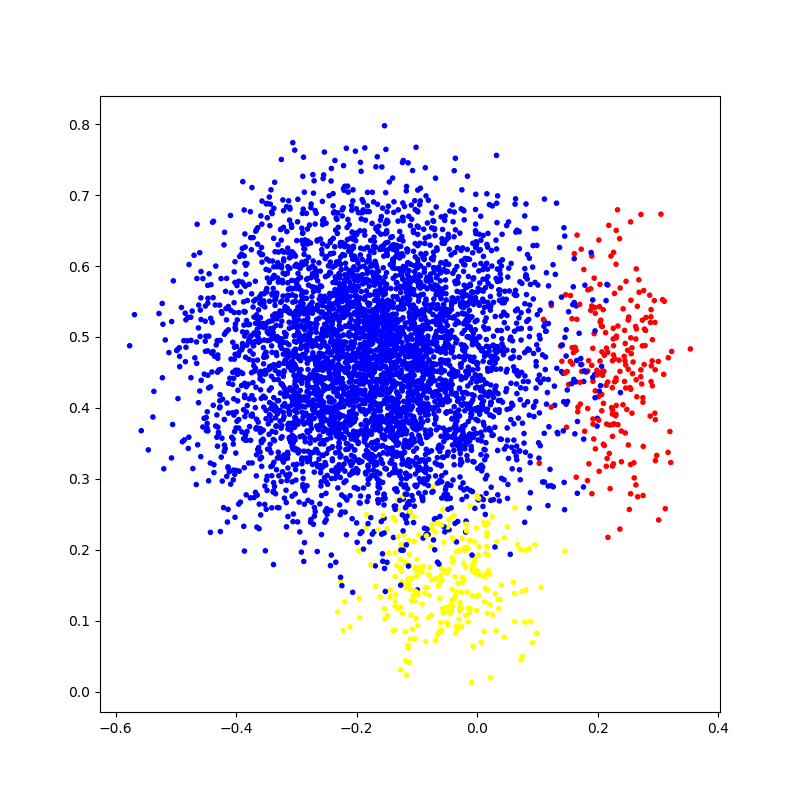} 
    \end{minipage}
    \caption{Two dimension visualization of examples from three classes generated by the new generator.}
     \label{newgener}
\end{figure}

For both generators it is possible to model all planed scenarios with considered parameters. Note that here we specially designed a new generator to better investigate the effect of overlapping between classes, in particular between minority classes. Below we present two simple 2-D  illustrations of 
one shot  of the example distribution from three classes generated by both generators. Firstly the result from the old generator  is presented in  Figure \ref{oldgen}, where colors were used to distinguish different classes and their borderline zones. While the 2-D illustration of the example distribution coming from the new generators is presented in Figure \ref{newgener}, and the reader can easily notice differences, in particular in the old generator the majority classes are always modeled as a single class, which surrounds the minority classes.

All stationary streams consist of 200,000 generated examples, while drifting streams have 250,000 examples. They are modeled with 5 attributes\footnote{Note that we are not modeling highly dimensional streams. In our opinion staying with 5 attributes is sufficient to model considered data difficulty factors as it was already done for static imbalanced data -- please compare experiments for static data such as \cite{LNS2017,Krysia_2016}, discussions on highdimensionality of learning problems \cite{jelonek1997feature}. Furthermore related works, such as \cite{DBLP:journals/tkde/DitzlerP13,WangMY15} are also done with a relatively small number of features.}. 

The single incremental drift is always started at 70 000 example and ends before 100 000 example. The classification performance  is  measured in every incoming 1000 examples. For the future presentation of results in next sections, we will show results for the following points: after the start of stream, the pre-drift (at 70,000 example), post-drift (100,000 where the lower decrease is usually observed in plots) and at the end of the stream. Unless stated otherwise, for drifting scenarios the minority classes are generated as single clusters of safe-type instances and all classes are balanced

\textit{Classifiers}: We chose four different online classifiers - the same as studied in \cite{brzezinski2021impact}\footnote{Recall that our study aims at detailed investigating the impact of selected factors on the classification of multi-class imbalanced stream, not at comparing many different classifiers such as \cite{aguiar2022survey}, so it is sufficient to select few representative classifiers only.}. Two of them are not specialised for dealing with imbalanced data streams: \textit{Online Bagging} (OB) ensemble \cite{Oza} and VFDT single classification tree \cite{vfdt}. The next ensembles are designed to deal with imbalanced streams: OOB which is an extension of OB with the global \textit{oversampling scheme} and UOB which is an extension of OB but with the \textit{undersampling scheme} \cite{WangMY15}. In all ensembles 15 VFDT are always used as component classifiers. All their parameters are default ones and the same as in \cite{brzezinski2021impact}.

\textit{Evaluation measures}: To be consistent with \cite{brzezinski2021impact} we evaluated classifiers with local accuracies of each class (recalls) and their aggregation to G-mean (so the geometric mean of recalls of all classes).  Such measures are also used in other, earlier studies, see e.g. \cite{imb-chapter-book,Korycki2021,WangMY15} and consistent with more general discussions of usefulness of various measures for evaluation classifiers in case of imbalanced data -- see \cite{brzezinski2018visual} which justifies why G-mean should be used instead F1-score or MCC for stronger imbalances between classes.

Implementations of generators and examined classifiers were written in Java for the MOA data stream framework \cite{moa}. Also all experiments were conducted in MOA command line mode similarly to ones described in \cite{brzezinski2021impact}.

\section{Experiments with single drifts or data difficulty factors}
\label{sec:experiments:single}

Firstly we analysed each single factor isolated from others in a similar way as in \cite{brzezinski2021impact}. In further subsections we will abbreviate the old generator by \textbf{O} and new one by \textbf{N}; in the names of data streams the single factors will denoted by number referring to their values, e.g. imb\_0.01\_0.01 is a balanced data stream with the static imbalance 1\% for two minority class and one majority class.

\subsection{Imbalance between classes}

The fully balanced stationary stream without any difficulty factors was quite easy to learn, where most classifiers achieved $\approx$ 0.99 G-mean and class recalls. The average classifier performance on \textit{stationary imbalanced streams} with minority class ratios 10\%, 5\%, and 3\% were nearly the same. Figures and detailed results are omitted due to page limits - the reader is referred to \cite{appendixLipska} to see some of these plots. Below in  Figures \ref{figure:imb11GN} -- \ref{figure:imb11GO} we present plots for the scenario of datastreams generated by the old generators containing two minority class with ratios 10\% and the one majority class with 80\% cardinality\footnote{The majority class in Figures \ref{figure:imb11GN} -- \ref{figure:imb11GO}  is denoted as class 0, while minority classes are class 1 and class 2.}.

\begin{figure}[h]
 \centering
    \begin{minipage}{0.45\textwidth}
        \centering
        \includegraphics[width=0.9\textwidth]{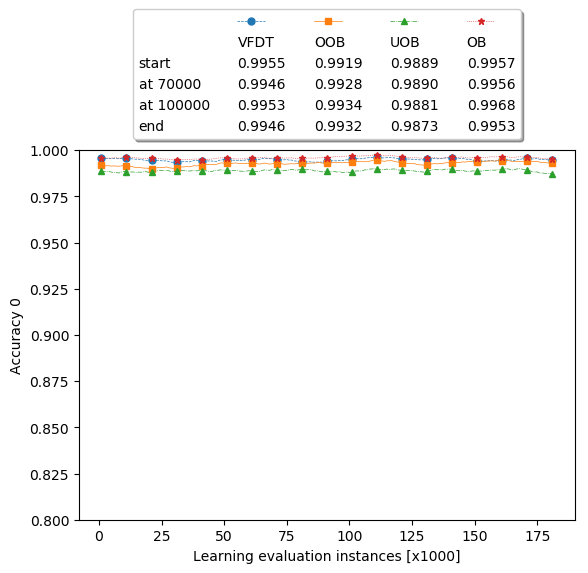} 
        \caption{Imb\_10\_10, Recall class 0}
        \label{figure:imb11GN}
    \end{minipage}\hfill
    \begin{minipage}{0.45\textwidth}
        \centering
        \includegraphics[width=0.9\textwidth]{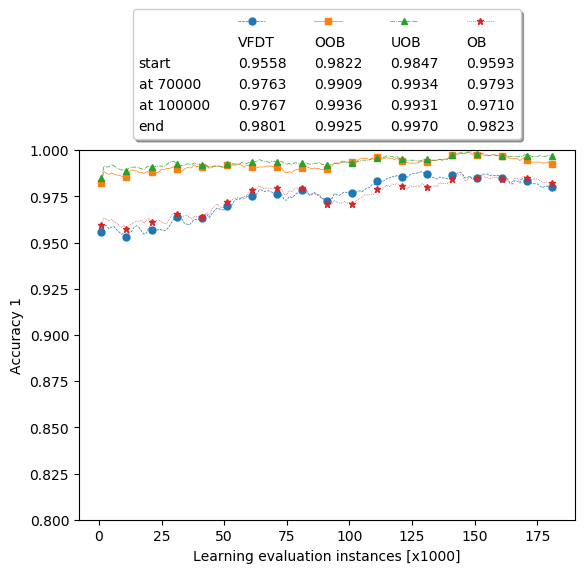} 
\caption{Imb\_10\_10, Recall class 1}
 \label{figure:imb11GO}
    \end{minipage}

\end{figure}
\begin{figure}[h]
 \centering
    \begin{minipage}{0.45\textwidth}
        \centering
        \includegraphics[width=0.9\textwidth]{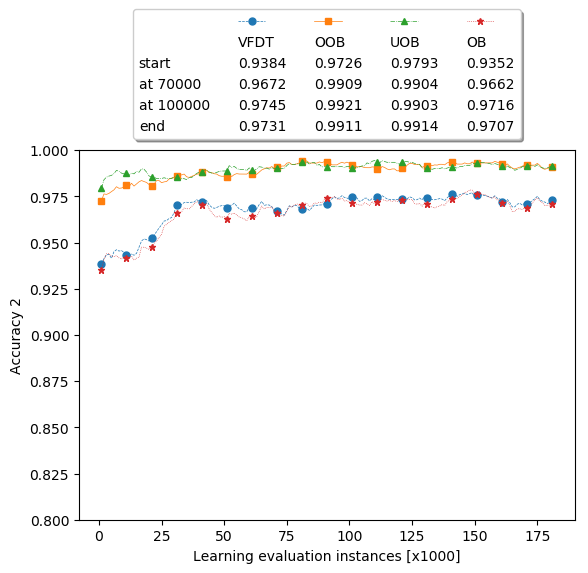} 
        \caption{Imb\_10\_10, Recall class 2}
        \label{figure:imb11GN}
    \end{minipage}\hfill
    \begin{minipage}{0.45\textwidth}
        \centering
        \includegraphics[width=0.9\textwidth]{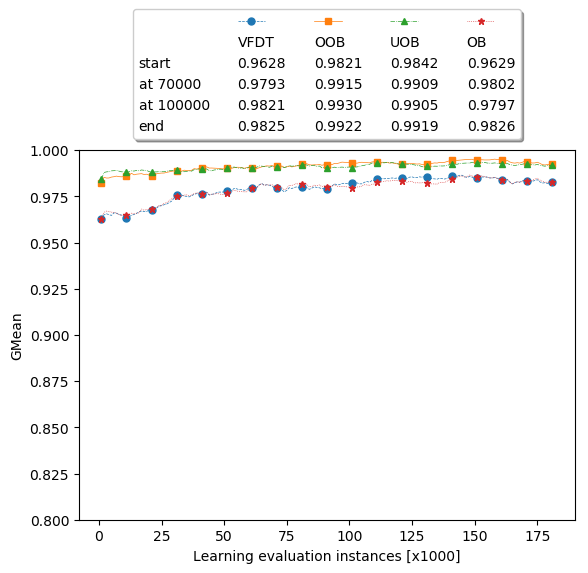} 
\caption{Imb\_10\_10, O, G-Mean}
 \label{figure:imb11GO}
    \end{minipage}
\end{figure}

For all the aforementioned minority class ratios, classifier performance plots looked very similar --- performance values rise fairly quickly up to a certain level and remain stable until the end of the stream. For instance, for imb\_0.03\_0.03 and the old generator  G-means are (VFDT 0.9366; OOB 0.9866; UOB 0.9928; OB 0.9356) while for \textbf{N} (VFDT 0.9367; OOB 0.9519; UOB 0.9562; OB 0.9447). Even for imb\_0.01\_0.01 G-means are similar: 2-3\% decreases for OOB and UOB, with stronger decrease for VFDT, e.g. for \textbf{O} (0.7452) and OB 0.6792.  The data streams from the old generator are always slightly more difficult than for the new one for stronger imbalance 1\%   while being easier for smaller imbalance ratios ($ \geq 3\%$); 

Increasing the number of the classes is more influential for the data streams coming  from the new generator, e.g.  compare imb\_0.03\_0.03\_0.03,   \textbf{N} (VFDT 0.8733; OOB 0.8876; UOB 0.8896; OB 0.8801) against imb\_0.03\_0.03. While for the older generator the increase of the number of class is not so visible. Considering scenarios with different class imbalance ratio, such as e.g.  imb\_0.03\_0.06\_0.12. we did not observe big differences of G-mean to the configuration with the same ratios, although usually the local Recall for the second class was slightly worse. 

Comparing the classifiers we can establish the following ranking UOB $\succeq$ OOB $\succ$ OB $\approx$ VFDT.  For imbalance ratios ($ \geq 5\%$) the differences between them are smaller while the differences increase for the stronger imbalances. Again these difference are smaller for \textbf{N} and are more visible for \textbf{O}. Usually UOB is the best classifier in the majority of scenarios. 

For \textit{imbalance ratio drifts} in data streams we did not notice big differences to the static equivalents due to the values of  measures. For stronger drifts (1\%, 3\% or 5\%) a small difference in G-mean (0.01-0.03) occurs mainly for VFDT and OOB but the final values are similar or even slightly higher than for the static streams. OOB and UOB generally are not affected by these drifts. For instance for dimb\_0.01\_0.01; G-mean values in moments $pre \rightarrow postdrift$ are the following \textbf{O} (VFDT 0.9901 $\rightarrow$  0.9852; OOB 0.9933 $\rightarrow$ 0.9915 ; UOB 0.9933 $\rightarrow$  0.9940; OB 0.9914 $\rightarrow$  0.9874) while for \textbf{N} (VFDT 0.9681 $\rightarrow$ 0.9544 ; OOB 0.9714 $\rightarrow$  0.9602; UOB 0.9703 $\rightarrow$  0.9606; OB 0.9704 $\rightarrow$  0.9566). For studied configuration final values of G-means in drifting streams are slightly higher ($ \geq 3\%$) than for static imbalanced stream (usually approx. 0.03) for the both generators. The increasing of the number of classes led to similar observations.   

\textit{A comparison with the binary imbalanced streams}: The obtained results for the old generator are quite similar except classes with imbalance ratios $1\%$, where for multiple classes the values of measures are better than for previous binary classes. The ranking of examined classifiers is also the same. What is new for multiple minority classes. There is no strong decrease of classifier performance for imbalanced ratio $1\%$ as it was observed for analogous drifts in the binary streams.

\subsection{Borderline and rare types of examples}

The G-mean results for all static configurations with different proportions of the types of examples in minority classes are presented in Table \ref{tab:statictypes}. The minority classes imbalance ratios are set to 10\% (following the previous section it is not an influential ratio).

\begin{table}[hbt]
  \begin{center}
  \caption{Comparison of the influence of various static levels of borderline or rare types of minority examples on classifier performance. G-mean averaged over entire streams for a given classifier on a given stream.}
  \label{tab:statictypes}
  \begin{tabular}{|l|llllllll|}
  \toprule
   datastream  & \multicolumn{4}{|c|}{Old generator} & \multicolumn{4}{|c|}{New generator} \\
   & VFDT & OOB & UOB & OB & VFDT & OOB & UOB & OB \\
  \midrule
  \multicolumn{9}{|c|}{Borderline : percentage of example types in minority classes} \\
  20 20 & 0.8600 & 0.9100 & 0.8956	 & 0.8606 & 0.9558 & 0.9681 & 0.9701 & 0.9617 \\
40 40 & 0.7533	& 0.8943 &	0.8623	& 0.7474 &	0.8357 & 0.8773 & 0.8675 & 0.8346 \\ 
60 60 & 0.6978	& 0.8950 &	0.8639	& 0.6810 & 0.665 & 0.7692 & 0.7752 & 0.6684 \\
80 80 & 0.6147 & 0.8922 & 0.8424	& 0.6006 & 0.1467 & 0.5367 & 0.5842 & 0.0902 \\
100 100 & 0.4326 & 0.8954 & 0.8140 & 0.4370 & 0.0 & 0.1903 & 0.4834 & 0.0 \\ \hline
  20 20 20 & 0.8588	& 0.9071 &	0.8939	& 0.8515 & 0.9022 & 0.9069 & 0.9102 & 0.9054 \\
40 40 40 & 0.7780	& 0.8669 & 0.8523	& 0.7566 & 0.7372 & 0.7710 & 0.7764 & 0.7370 \\
60 60 60 & 0.7287	& 0.8554 & 0.8445	& 0.7106 & 0.5809 & 0.6732 & 0.6787 & 0.5747 \\
80 80 80 & 0.6921	& 0.8485 & 0.8227	& 0.6841 & 0.1540 & 0.4450 & 0.4704 & 0.0949 \\
100 100 100 &	0.5920	& 0.8384 &	0.7934	& 0.5679 & 0.0 & 0.1578 & 0.3737 & 0.0 \\ \hline
 \multicolumn{9}{|c|}{Rare cases : percentage of example types in minority classes} \\
 20 20 & 0.8442 & 0.8492 & 0.8255 & 0.8296 & 0.8312 & 0.8464 & 0.8499 & 0.8374 \\
40 40 & 0.6826 & 0.6990	& 0.6770 & 0.6855 & 0.6963 & 0.7680 & 0.7400 & 0.6949 \\
60 60 & 0.5177 & 0.5314	& 0.5275 & 0.5131 & 0.5365 & 0.7058 & 0.6567 & 0.5275 \\
80 80 & 0.2599 & 0.3329	& 0.4351 & 0.2559 & 0.3559 & 0.6186 & 0.5991 & 0.3276 \\ 
100 100 &  0.0052 & 0.0095	& 0.3502 & 0.0 & 0.1551 & 0.5557 & 0.5614 & 0.0350 \\ \hline
 20 20 20 & 0.8163 & 0.8328	& 0.7882 & 0.8133	& 0.7763 & 0.7762 & 0.7781 & 0.7793 \\
40 40 40 & 0.6639 & 0.6688	& 0.6093 & 0.6647 & 0.6287 & 0.6889 & 0.6516 & 0.6300 \\
60 60 60 & 0.4634	& 0.4910 & 0.4723 & 0.4811 &  0.4601 & 0.5988 & 0.5426 & 0.4651 \\
80 80 80 & 0.1999	& 0.2973 & 0.1505 & 0.0949	& 0.2789 & 0.4877 & 0.4715 & 0.2595 \\
100 100 100 &  0.0046 & 0.002 & 0.0173	& 0.0 & 0.0985 & 0.3994 & 0.3847 & 0.1175 \\ 
  \bottomrule
  \end{tabular}
  \end{center}
\end{table}

For static \textbf{borderline} examples one can easily notice the big decrease of G-mean for all classifiers, in particular for the older generator. Increasing the presence of the borderline examples in the minority classes causes a very marked decrease for VFDT and OB while being smaller for OOB and UOB (but they loose 0.1 for \textbf{O}, while much more for \textbf{N}). Increasing the number of minority classes does not change these trends. Moreover we notice that the Recalls of the second minority class, in the data streams from the new generator are lower than other minority classes. It may be caused that its neighborhood classes also increase their overlapping with the higher number of examples.  

\textit{Borderline ratio drift}: For both generators a decrease in G-mean was observed after the drift and classifiers did not recover to pre-drift levels. OB and VFDT performance is clearly worse than UOB and OOB.  Some exemplary results are presented in Table \ref{tab:driftypes}.  

\begin{table}[hbt]
  \begin{center}
  \caption{Comparison of the influence of selected drifts of borderline or rare types of minority examples on classifier performance. G-mean  calculated in three moments of the stream.}
  \label{tab:driftypes}
  \begin{tabular}{|l|llllllll|}
  \toprule
   data stream  & \multicolumn{4}{|c|}{Old generator} & \multicolumn{4}{|c|}{New generator} \\
   moments & VFDT & OOB & UOB & OB & VFDT & OOB & UOB & OB \\
  \midrule
  \multicolumn{9}{|c|}{Borderline : percentage of example types in minority classes} \\
  start & 0.9628 &	0.9826 & 0.9839	& 0.9612 & 0.9190 & 0.9191 & 0.9192 & 0.9193 \\ \hline
  d\_20\_20 pre & 0.9347 & 0.9459 & 0.9456 &	0.9345	& 0.9185 & 0.9264 & 0.9239 & 0.9212 \\
post & 0.8501 & 0.8816 & 0.8880 & 0.8467 &	0.9025 & 0.9119 & 0.9052 & 0.9054 \\
end	& 0.8528 & 0.9064 & 0.8880	& 0.8483 & 0.9065 & 0.9192 & 0.9157 & 0.9119 \\ \hline
 d\_40\_40 pre & 0.8894	& 0.9041 & 0.8991 &	 0.8882	& 0.8938 & 0.9087 & 0.9059 & 0.8982 \\
post & 0.7243 & 0.8269 & 0.8365 & 0.7067 &	0.8077 & 0.8515 & 0.8554 & 0.8147 \\
end  & 0.7818 & 0.8934 & 0.8479 & 0.7651 & 0.8427 & 0.8700 & 0.8636 & 0.8476 \\ \hline
d\_60\_60 pre &	0.8482	& 0.8700 & 0.8780 & 0.9492	& 0.9170 & 0.9410 & 0.9454 & 0.9221 \\
post & 0.6273 & 0.7733 & 0.8371	& 0.6085 & 0.6956 & 0.8101 & 0.7979 &0.7006 \\ 
end	& 0.6749 & 0.8969 & 0.8797	& 0.6653 & 0.7904 & 0.8536 & 0.8532 & 0.7758 \\ \hline  
d\_100\_100 pre & 0.7519	& 0.7982 & 0.8128 & 0.7506 & 0.6735 & 0.7731 & 0.7759	& 0.6731 \\ 
post & 0.4407 & 0.7657 & 0.8159 &  0.3942 &	0.0115 & 0.3410 & 0.1057 & 0.0087 \\
end & 0.4815 & 0.8906 & 0.8451 &  0.3506 & 0.0122 & 0.5103 & 0.5640 & 0.0029  \\ \hline
 \multicolumn{9}{|c|}{Rare cases : percentage of example types in minority classes} \\
start  &0.9628 & 0.9826 &  0.9839 & 0.9612 & 0.9190 & 0.9281 & 0.9313 & 0.9294 \\ \hline
d\_20\_20 pre	& 0.9274 & 0.9409	& 0.9412 &	0.9278 & 0.8806 & 0.8904 & 0.8864 & 0.8882 \\
post & 0.8246 &0.8427 & 0.8328	& 0.8273 & 0.7868 & 0.7918 & 0.7932 & 0.7888 \\
end	& 0.8086	& 0.8346	& 0.8275	& 0.8178 & 0.7890 & 0.7950 & 0.7966	& 0.7989 \\ \hline
d\_40\_40  pre	& 0.885 &  0.8959 & 0.8956 & 0.8882 & 0.8837 & 0.8957 & 0.8941 & 0.8879 \\
post & 0.6913 & 0.7022	& 0.7005 & 0.6918 & 0.6955 & 0.7256 & 0.7177 & 0.6957 \\
end	& 0.6842 & 0.6961 & 0.6967 & 0.6854	& 0.6895 & 0.7723 & 0.7755 & 0.6913 \\ \hline
d\_60\_60 pre & 0.8370 & 0.8469	& 0.8479 & 0.8348 & 0.8394 & 0.8553 & 0.8491 &0.8403 \\
 post &0.5335	& 0.5444	& 0.5404 & 0.5336 & 0.5285 & 0.6396 & 0.6103 & 0.5300 \\
end	& 0.5248 & 0.5385 & 0.5361& 0.5247	& 0.5338 & 0.7058 & 0.7094 & 0.5328 \\
  \bottomrule
  \end{tabular}
  \end{center}
\end{table}

For the old  generator all classifiers achieve similar levels of G-Mean after the drift  as for the static borderline ratio  except ratio 100\% where values are higher. The increasing the number of minority classes does not affect performance. For the new generator all classifiers achieve a bit higher G-mean than for the static counterpart, again except ratio 100\% (which was completely deteriorated for the static streams). Moreover for the new version of the generator, an increase of the number of minority classes decreases G-means (by 0.05-0.1). Comparing classifiers OB and VFDT performance is clearly worse than UOB and OOB. On contrary to the imbalance ratios, now OOB is slightly better than UOB.

\textit{A comparison with the binary imbalanced streams}: For multiple minority classes the decreases of the measures are higher than for binary imbalanced streams. It concerns both static and drifting scenarios. For instance, for binary class, static streams and percentage of  borderline examples 20\%,  classifiers' performance was $\approx$ 0.95 while for multiple classes OOB, UOB are $\approx$ 0.90 and OB, VFDT are $\approx$ 0.85. Percentages of borderline examples closer to 100\% completely deteriorate the classifiers' performance. Furthermore their reactions to the borderline drifts are also stronger with the worse recovery than for the binary streams.


\noindent \textbf{Rare examples}: 
\textit{Stationary streams} - the influence of an increasing the percentage of rare types of minority examples is similar to static borderline examples, i.e. they decrease the performance of all classifiers although values of all measures are clearly lower; see Table \ref{tab:statictypes}. For example for Bor\_40\_40 all classifiers have G-Mean values higher than $\approx$ 0.75 and for Rare\_40\_40 all classifiers have G-Mean value around 0.68. For higher percentages of rare examples VFDT and OB are unable to learn classes (for borderline case it appeared only for \textbf{N} and 100\%) while OOB and UOB perform better  (although with worse values than for similar configurations of borderline examples).   For the new generator and for the percentage of rare examples at least 40\% classifiers (OB, VFDT) are clearly separated from specialized methods (OOB,UOB). For the old version of generator the number of minority classes decreases G-Mean values for at least 40\% rare percentage. For the new generator influence of the number of minority classes on G-Mean values increases with rising rare ratio (G-Mean value decreases by 4-15\% for 3 classes scenarios). While comparing different classifiers,  OOB has G-Mean values slightly  higher (around 3\%) than UOB.

In general rare examples are clearly more difficult than borderline examples and both are worse than highly imbalanced streams. For example values of G-Mean for Rare\_40\_40 are: $\approx$ 0.67 for OB and VFDT, 0.68 for UOB and 0.69 for OOB, while for Imb\_10\_10 --  they are 0.98 for OB and VFDT and 0.99 for UOB and OOB.

\begin{figure}[h]
 \centering
    \begin{minipage}{0.45\textwidth}
        \centering
        \includegraphics[width=0.8\textwidth]{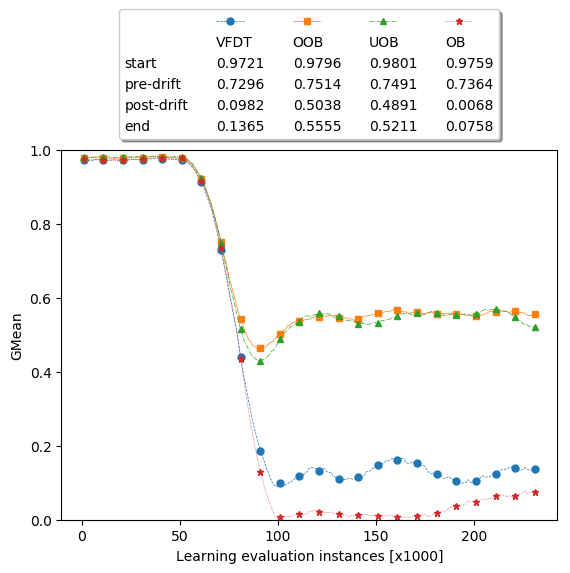} 
        \caption{G-mean plot  for data stream  Rare drift\_100\_100 and N generator}
        \label{figure:rare}
    \end{minipage}\hfill
    \begin{minipage}{0.45\textwidth}
        \centering
        \includegraphics[width=0.8\textwidth]{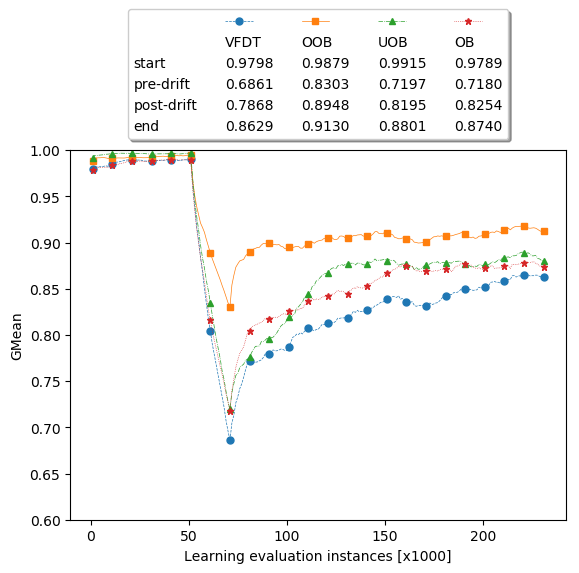} 
\caption{G-mean plot  for Split\_5\_5\  and O generator}
\label{figure:split}
    \end{minipage}
\end{figure}

\textit{Local drifts of rare types of minority examples}: -- some results are shown in Table \ref{tab:driftypes} and in Figure \ref{figure:rare}. For both generators a decrease of G-Mean values is observed after the drift and classifiers do not return to pre-drift levels. In most cases there is not even a small increase in G-Mean. All classifiers after drift finally achieve similar levels of G-Mean as for static rare ratio experiment - for instance compare values of d\_rare\_40\_40 and rare\_40\_40 values. For the old version of generator the number of minority classes decreases G-Mean values for rare ratio after drift. For rare ratios 20\%, 40\%, 60\%, 80\% decrease is 2\%, 2\%, 5\%, 5-11\% respectively. For the new generator and  for rare ratio after drift equal at least 40\%  classifiers (OB, VFDT) are clearly separated from specialized methods (OOB,UOB) -- they have lower G-Mean values -- while for the old generator similar differences occur for the higher percentages of rare examples drifts. 


\textit{A comparison with the binary imbalanced streams}: There is some similarity between static binary and multiple class problems. For the percentage of 20\% rare examples G-Mean values for binary problem are around 0.85 while for multiple minority class streams around 0.83-84. For drifts of rare examples the increasing the number of minority classes additionally decreases of G-mean values after drifts.  For instance, for rare ratio after drift 20\% G-Mean values for all classifiers  for binary problem are $\approx$ 0.88,  while for 2 minority classes $\approx$  0.85 and for three minority classes $\approx$  0.83. Finally it is clearly visible that for both generators rare example drifts are more difficult than binary ones

\subsection{Class composition splits and move}

\noindent \textit{Experiments with minority classes moves}: Following \cite{brzezinski2021impact} we model configurations of decomposing each minority class into 3, 5 and 7 sub-clusters and, then, model their moves into new positions (randomly chosen but in such a way that they not intersect). For both generators the cluster movement drift causes G-Mean decrease, although decrease is not permanent and classifiers recover -- for some classifiers even to higher values than in pre-difts. Values for \textbf{N} generator are usually slightly higher than for \textbf{O}. Both decrease and recovery are visible in contrast to previously discussed drift scenarios (imbalance ratio, instance type).  For both generators adding third minority class causes G-Mean decrease by 0.03 -- 0.05. There are some difference between classifiers:  OOB has the highest G-Mean values and the lowest decrease after drift.

\begin{table}[hbt]
  \begin{center}
  \caption{Drifts of minority classes splits into sub-clusters on classifier performance; and one representative move drift. G-mean  calculated in four moments of the stream - start, pre-drift, post-drift and the end}
\label{tab:driftysplits}
  \begin{tabular}{|l|llllllll|}
  \toprule
   data stream  & \multicolumn{4}{|c|}{Old generator} & \multicolumn{4}{|c|}{New generator} \\
   moments & VFDT & OOB & UOB & OB & VFDT & OOB & UOB & OB \\
  \midrule
 \multicolumn{9}{|c|}{Split each minority class into sub-clusters} \\  
 3\_3 start & 0.9725  &	0.9875	& 0.9876 & 0.9684 & 0.9836 & 0.9881 & 0.9921 & 0.9840 \\
pre	 & 0.7314 & 0.8494 & 0.8135	& 0.700	& 0.8836 & 0.9496 & 0.8773 & 0.8894 \\ 
post & 0.8400 & 0.9152 & 0.9016	& 0.8565 & 0.9474 & 0.9764 & 0.9608 & 0.9607 \\
end	& 0.8936 & 0.9240 & 0.9168 & 0.8981 & 0.9660 & 0.9794 & 0.9760 & 0.9666 \\ \hline
5\_5 start & 0.9798  & 0.9879 & 0.9915 & 0.9789 & 0.9893 & 0.9928 & 0.9939 & 0.9908 \\
pre	& 0.6861 & 0.8303 & 0.7197	& 0.7180 & 0.8722 & 0.9361 & 0.8851 & 0.8817 \\
post & 0.7868 & 0.8948 & 0.8195	& 0.8254 & 0.9382 & 0.9691 & 0.9474 & 0.9523 \\
end	& 0.8629 & 0.9123 & 0.8801	& 0.8740 & 0.9512 & 0.9691 & 0.9528 & 0.9583 \\ \hline
 \multicolumn{9}{|c|}{Move sub-cluster drift} \\
5\_5 start & 0.668	& 0.8847 & 0.8523 &	0.6553 & 0.7799 & 0.7935 & 0.7803 & 0.7875 \\
pre	& 0.7254 & 0.8103 & 0.8018	& 0.7368 &	0.8586 & 0.9098 & 0.8548 & 0.8633 \\
post & 0.8505 & 0.9510 & 0.8766	& 0.8780 &	0.8954 & 0.9150 & 0.8561 & 0.9003 \\
end	& 0.9139 & 0.9766 & 0.9299	& 0.9525 & 	0.9308 & 0.9487 & 0.8772 & 0.9332 \\
  \bottomrule
  \end{tabular}
  \end{center}
\end{table}

\noindent \textit{Experiments with minority cluster splits}:  The class decomposition drifts are modeled as starting with single cluster shape for each minority class and then splitting it into 3, 5 or 7 non-overlapping sub-clusters. Some results are present in Table \ref{tab:driftysplits}. For both versions of generators, the class split causes decrease of G-Mean values. What is more, classifiers do not return to pre-drift levels -- see an example in Figure \ref{figure:split}. Increasing the number of splits decreases values of G-mean after drifts for the old generator while is not so influential for the new one. For the old version of generator adding third minority class slightly increases G-Mean values by 1-3\% for OOB and UOB and by 3-5\% for OB and VFDT for Split\_3\_3 and Split\_5\_5. For the new generator the adding more minority class decreases G-Mean by 1-3\%. Comparing decreases with previous drifts, rare example drift and then borderline drift are more difficult than the split drift. For example for Bord\_20\_20 all classifiers have G-Mean values lower than 0.91 and for Rared\_20\_20 -- lower than 0.86 while for all split drifts all classifiers have G-Mean values above 0.93. It was expected because rare and borderline drifts increase the number of minority instances in majority class significantly more than split drift, which leaves the overlap ratio unchanged. For both generators the highest scores considering G-Mean has OOB. Moreover, for this classifier decrease after drift is the lowest: around 0.03-0.13 less than the second best for the old generator  and 0.03-0.05 less for the new one (UOB is usually the second in the order).

\textit{A comparison with the binary imbalanced streams}: Similarly now G-Mean values after drift decreases and classifiers do not return to pre-drift levels. For binary problem G-Mean values decreases from 0.98 to $\approx$  0.96 for all classifiers. For multiple minority class problem G-Mean values decreases from  to 0.86-0.92 and classifiers have different G-Mean values. It is worth noting that for binary problem G-Mean decrease after drift is around 10-15\% and it is almost the same for all classifiers. For multiple minority class problem OOB has the lowest decrease of all classifiers which which was not so clear for the binary streams.

\section{Experiments with complex scenarios}

The interactions between data difficulty factors and drifts is examined to see how much such conjunctions additionally decrease the classifiers' performance.  Here we follow some combined scenarios from \cite{brzezinski2021impact}

Firstly we combine the imbalance ratios with all other factors. Generally they do not change the trends for these factors. Unlike the binary imbalanced streams \cite{brzezinski2021impact} (where the minority class ratios 1\% -- 3\% amplified decreases of the measures) here only the highest ratio 1\% led to very small additional decreases and did not change shapes of plots.  The more interesting results are for combinations of other factors. We have chosen the split the minority classes into 5 sub-clusters (recall that 3 splits give similar results)  and the drift of instance type examples of borderline or rare  with two variants of percentages 20\% and 60\% (the second is more influential and plausible with respect to occur in streams  in contrast to the higher values).

\begin{figure}[h]
 \centering
    \begin{minipage}{0.45\textwidth}
        \centering
        \includegraphics[width=0.8\textwidth]{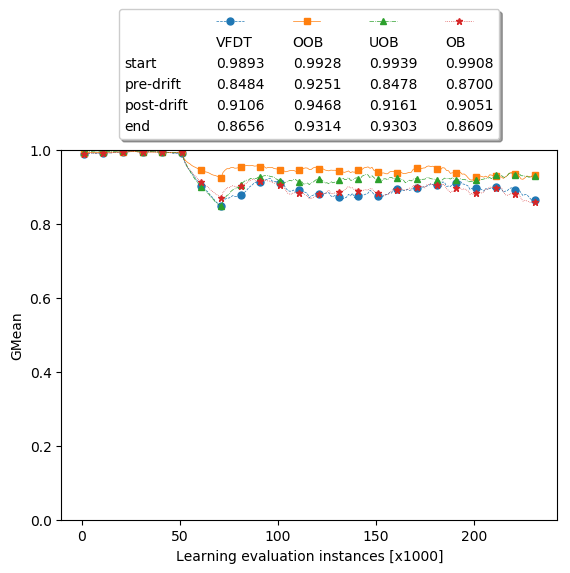} 
        \caption{G-mean plot  for data stream  Split\_5\_5\_Imbd\_1\_1\_Bord\_60\_60 and N generator}
        \label{figure:splitborimb6060GN}
    \end{minipage}\hfill
    \begin{minipage}{0.45\textwidth}
        \centering
        \includegraphics[width=0.8\textwidth]{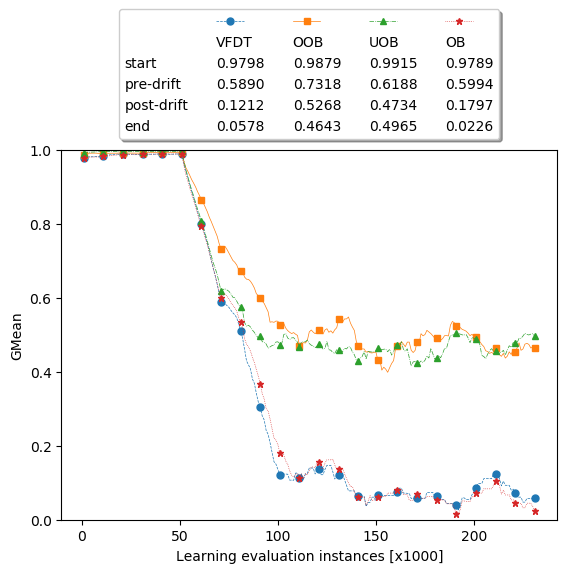} 
\caption{G-mean plot  for Split\_5\_5\_Imbd\_1\_1\_Bord\_60\_60, data stream and O generator}
\label{figure:splitborimb6060GO}
    \end{minipage}
\end{figure}

\noindent \textbf{Cluster splitting and borderline drift}.   For the older generator this scenario led to lower values of G-mean than for borderline drift alone by 0.07 -- 0.18 depending on the classifier. On the other hand for the new generator differences are smaller $\approx$ 0.02 (Compare Figures \ref{figure:splitborimb6060GN} and \ref{figure:splitborimb6060GO}). For a version with the third minority class we observed additional decrease of G-mean values by 0.03-0.04, while for the new generator it does not influence the results. There is also a clear difference between the better performing classifiers (UOB, OOB) and worse (VFDT,OB) for \textbf{O} generator while much less for \textbf{N}. 

\textit{Comparison with binary streams}: For binary problem, the split drift dominated the borderline drift in contrast to multiple minority class problem, where the result is a combination of both phenomena. G-Mean values decrease about 0.12 comparing to split experiment for binary problem. Furthermore values of G-Mean are 0.07-0.18 lower than for borderline ratio drift. Generally, G-Mean decrease for multiple minority problem is 0.05-0.10 stronger than for the binary stream.

\begin{figure}[h]
 \centering
    \begin{minipage}{0.45\textwidth}
        \centering
        \includegraphics[width=0.8\textwidth]{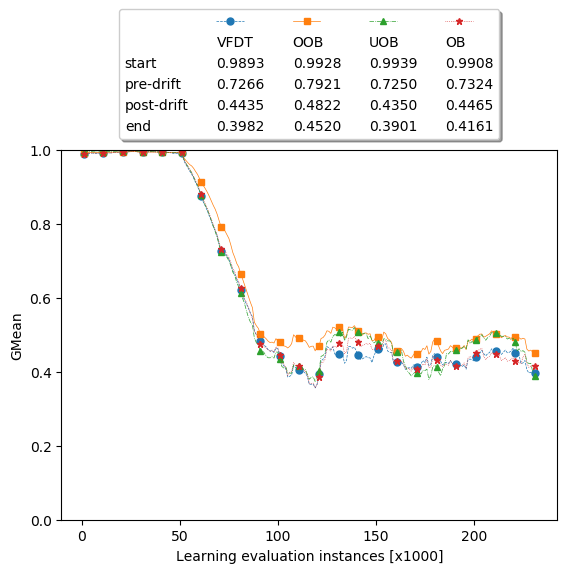} 
        \caption{G-mean plot of all classifiers for Split\_5\_5\_Imbd\_1\_1\_Rared\_60\_60 data stream and N generator}
        \label{figure:splitrareimb6060GN}
    \end{minipage}\hfill
    \begin{minipage}{0.45\textwidth}
        \centering
        \includegraphics[width=0.8\textwidth]{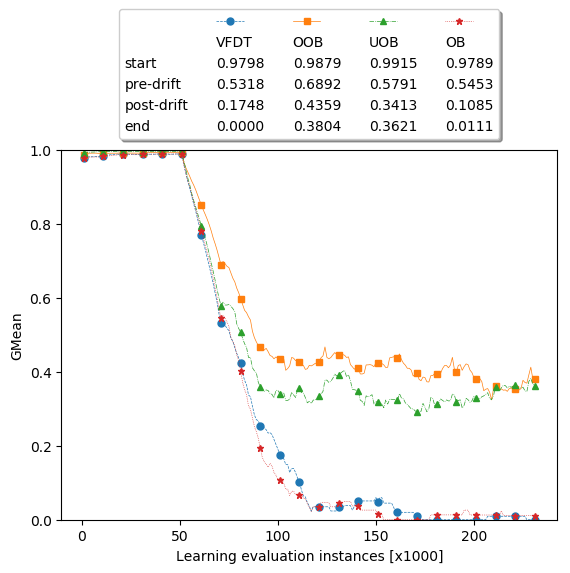} 
\caption{G-mean plot  for Split\_5\_5\_Imbd\_1\_1\_Rared\_60\_60 data stream and O generator}
\label{figure:splitrareimb6060GO}
    \end{minipage}
\end{figure}


\noindent \textbf{Cluster splitting and rare drift}. This combination also strongly decreases the values of measures with respect to single factors. For instance,  G-Mean values after drift are lower than for rare drift by 0.03-0.08 for OOB, 0.05-0.16 for OB and VFDT and 0.09-0.20 for UOB for \textbf{O} generator while less for \textbf{N} generator (compare Figures \ref{figure:splitrareimb6060GN} vs. \ref{figure:splitrareimb6060GO}). Moreover, the plots for classifiers are similar to  the plots for rare drift, which means that for above 40\% rare ratio classifiers  (VFDT,OB) are clearly separated from better specialized (OOB, UOB) - see Figure \ref{figure:splitrareimb6060GO}. Finally OOB is the best classifier in this ranking. 

\textit{Comparison with binary streams}: Looking at plots the split drift seems to be dominated by the rare drift. Similarly to the binary streams, now the composition of split drift with rare drift is more difficult than the split drift with borderline drift.  An adding more classes makes the problem more difficult.

Then, we considered the  adding the imbalanced drift 1\% to these combined scenarios. These are the most difficult streams. For both generators and the variants of borderline drift it decreases G-Mean values for all classifiers, i.e.  for the new generator by 0.01-0.09 and for the old one by: 0.05-0.26 for OOB, 0.09-020 for UOB 0.38-0.54 for OB and VFDT (which is stronger than for two combined elements). The similar amplifications of measures' decreases also occur for combining cluster splitting, rare drift and imbalance drift. Here we observed the strongest decreases for all classifiers. Moreover, an increasing of the number of minority classes decreases G-mean and Recalls (more for \textbf{O} generator). 

\textit{Comparison with binary streams}: The scenario with the borderline drift was not so difficult as for multiple classes. Previously UOB was the best performing classifiers while it is now overtaken by OOB. For multiple classes the stream with rare example drift is more difficult than for binary streams and it is more clearly visible that VFDT and OB are almost unable to learn for the higher rare ratio. 

To sum up our observations, generating combined factors and drifts in a multiple minority class version makes the problem more difficult than counterparts with binary classes.  

\section{Final remarks}

The conducted experiments with multi-class imbalanced synthetic data streams allowed us to identify the most influential factors (ranking in the order of their impact): presence of rare types of minority examples, borderline examples and split of the minority class into sub-clusters.  The drift of moving sub-clusters in each classes had  very little impact -- all classifiers well recovered after the drift to sufficiently high values. The imbalance ratios of minority classes were also not so influential. The classifiers specialized for imbalanced streams, i.e. OOB and UOB, performed really well in all of experiments with the static or drifting imbalance ratios.   Note that for imbalance ratio at least 10\% other classifiers (OB and VFDT) also performed quite well. However this partly surprising result is quite similar to the one observed for the binary problem, except for the imbalance ratio 1\%, which was unfeasible to be learnt by classifiers in the binary streams. In multiple classes streams such ratio only has some effect if its drift is added to complex scenarios. see e.g. Fig \ref{figure:splitrareimb6060GO}.

The combinations of three drifts: rare or borderline types of examples, the split of the minority classes into sub-cluster and their 1\% imbalances are the most difficult versions of streams. In particular the scenario with the rare examples is  the most difficult  one and it is more demanding than its binary class counterpart (the difference between the best classifiers for binary and the multiple minority class problem is $\approx$ 15\%). Looking at figures \ref{figure:splitrareimb6060GO}  or \ref{figure:splitborimb6060GO}  one can easily notice the high decreases of G-mean and the inability of all classifiers to recover their performance. Especially (VFDT, OB) lose the ability to recognize minority classes. Although comparing classifiers was not the aim of this work, the domination of specialized OOB and UOB ensembles in all experiments is clearly visible.

To sum up, the multiple minority class streams demonstrated to be more difficult than binary ones. Furthermore adding a third minority class generally worsened the G-mean values ($\approx$ 10\%). In particular, the recognition of the middle class for the new generator stream was weaker, which may be related to the greater overlap between the classes. Moreover, we observed that an increase of overlap between the minority classes and the majority class was more difficult than the variant with increasing overlap between minority classes only. This is quite consistent with similar observations from experiments with static multi-class data, see \cite{eswa2022}. 

Comparing the current results to the previous ones with binary classes, one should notice a much greater importance of class overlap, i.e. borderline types of examples. In addition, the significantly lower importance of strong imbalances - which is also particularly visible for the streams from the new generator. Furthermore, now OOB ensemble is more often slightly better than the UOB.

Future research  should focus on the identified difficulty factors more than just on the global class imbalances. As none of the analyzed classifiers were able to cope with rare or borderline drifts combined with minority class splits, the new proposed classifiers should  take into account these compositions of difficult factors. 

\noindent \textbf{Acknowledgements}: The research of Jerzy Stefanowski was partially supported by 0311/ SBAD/0726 PUT University grant.

\bibliographystyle{plain}
\bibliography{jmlr-sample}

\end{document}